\documentclass[journal,twoside,web]{ieeecolor}
\usepackage{tmi}
\makeatletter
\def\ps@titlepagestyle{\def\@oddfoot{}\def\@evenfoot{}\def\@oddhead{}\def\@evenhead{}}
\makeatother
\usepackage{amsmath,amssymb,amsfonts}
\usepackage{algorithmic}
\usepackage{graphicx}
\usepackage{textcomp}
\usepackage{url}
\usepackage[style=ieee, backend=biber, maxnames=6,
            url=false, doi=false, isbn=false, eprint=false]{biblatex}
\DeclareSourcemap{
  \maps[datatype=bibtex]{
    \map[overwrite=true]{
      \step[fieldsource=title, match=\regexp{DeepLab}, final]
      \step[fieldsource=shortjournal, final]
      \step[fieldset=journaltitle, origfieldval]
    }
  }
}
\AtEveryBibitem{%
  \clearfield{day}\clearfield{endday}%
  \clearname{editor}\clearname{editora}\clearname{editorb}%
  \clearlist{location}\clearlist{publisher}%
  \clearfield{series}\clearfield{note}\clearfield{issn}%
}
\DeclareUnicodeCharacter{2212}{-}                    
\DeclareUnicodeCharacter{2010}{-}                    
\DeclareUnicodeCharacter{00AD}{-}                    
\DeclareUnicodeCharacter{2009}{\,}                   
\DeclareUnicodeCharacter{202F}{\,}                   
\DeclareUnicodeCharacter{223C}{\ensuremath{\sim}}    
\DeclareUnicodeCharacter{00D7}{\ensuremath{\times}}  
\DeclareUnicodeCharacter{00B1}{\ensuremath{\pm}}     
\DeclareUnicodeCharacter{03BC}{\textmu}              
\DeclareUnicodeCharacter{FB01}{fi}                   
\DeclareUnicodeCharacter{FB02}{fl}                   
\def\BibTeX{{\rm B\kern-.05em{\sc i\kern-.025em b}\kern-.08em
    T\kern-.1667em\lower.7ex\hbox{E}\kern-.125emX}}
\makeatletter
\def\fnum@figure{{\color{subsectioncolor}\sffamily Fig.~\thefigure}}
\def\fnum@table{{\color{subsectioncolor}\sffamily TABLE~\thetable}}
\makeatother
\usepackage{etoolbox}
\makeatletter
\patchcmd{\@sect}{\noindent}{\noindent\set@color}{}{%
  \PackageWarning{tmi-fix}{could not patch \string\@sect\space for heading color}}
\makeatother

\begin{document}
\title{Beyond Isotropic Assumptions: Continuity-Constrained Segmentation and GPU Morphometry for Nanoscale GBM Analysis}
\author{Arash Fatehi, Robin Ebbestad, Linus Butt, Hans Blom, Sigrid Lundberg, Hannes Olauson, Hjalmar Brismar, David Unnersj\"{o}-Jess, Thomas Benzing, and Katarzyna Bozek
\thanks{A. Fatehi and K. Bozek are with the Institute for Biomedical Informatics,
Faculty of Medicine and University Hospital Cologne, University of Cologne,
Cologne, Germany.}
\thanks{A. Fatehi, D. Unnersj\"{o}-Jess, L. Butt, T. Benzing, and K. Bozek are with the
Center for Molecular Medicine Cologne (CMMC), University of Cologne, Faculty of
Medicine and University Hospital Cologne, Cologne, Germany.}
\thanks{D. Unnersj\"{o}-Jess, L. Butt, and T. Benzing are with Department II of Internal
Medicine, University of Cologne, Faculty of Medicine and University Hospital
Cologne, Cologne, Germany.}
\thanks{D. Unnersj\"{o}-Jess, L. Butt, T. Benzing, and K. Bozek are with the Cluster of
Excellence Cellular Stress Responses in Aging-Associated Diseases (CECAD),
University of Cologne, Faculty of Medicine and University Hospital Cologne,
Cologne, Germany.}
\thanks{D. Unnersj\"{o}-Jess is also with MedTechLabs, Karolinska University
Hospital, Solna, Sweden; the Division of Renal Medicine, Department of Clinical
Sciences, Intervention and Technology, Karolinska Institutet, Stockholm, Sweden;
and the Science for Life Laboratory, Department of Applied Physics, Royal
Institute of Technology (KTH), Solna, Sweden.}
\thanks{R. Ebbestad and H. Brismar are with the Science for Life Laboratory,
Department of Women's and Children's Health, Karolinska Institutet, Solna, Sweden.}
\thanks{H. Brismar is also with the Science for Life Laboratory, Department of
Applied Physics, Royal Institute of Technology (KTH), Solna, Sweden.}
\thanks{H. Blom is with the Science for Life Laboratory, Department of Applied
Physics, Royal Institute of Technology (KTH), Solna, Sweden, and MedTechLabs,
Karolinska University Hospital, Solna, Sweden.}
\thanks{S. Lundberg is with MedTechLabs, Karolinska University Hospital, Solna,
Sweden, and the Division of Nephrology, Department of Clinical Sciences, Karolinska
Institutet, Danderyd Hospital, Stockholm, Sweden.}
\thanks{H. Olauson is with the Division of Renal Medicine, Department of Clinical
Sciences, Intervention and Technology, Karolinska Institutet, Stockholm, Sweden.}}

\maketitle

\begin{abstract}
Confocal microscopy of optically cleared and swelled tissue resolves complex biological structures in 3D. Such acquisitions, however, are highly anisotropic: along the under-sampled axial direction the structure can appear discontinuous. This hampers isotropic reconstruction and automated quantitative analysis of morphologies in 3D. The usual approach upsamples the axial dimension to an isotropic volume before training a segmentation model. This requires dense annotations in the upsampled dimension, a prohibitive labeling burden for volumetric data. 
Here we present an end-to-end, GPU-accelerated framework that overcomes this limitation without the need for additional annotations. Our segmentation model is trained on the native acquisition volume. By adding random rotation to training patches, we leverage the higher resolution of the lateral plane to supply the missing axial information, and a \textit{z}-axis continuity loss keeps neighboring slices consistent. We adapt both a convolutional (3D U-Net) and a transformer (SwinUNETR) backbone. Overlapping patches are aggregated by Gaussian consensus, and point spread function (PSF)-corrected membrane thickness is computed by ray--surface intersection on the GPU. 
We apply our method to the segmentation and morphometric analysis of the glomerular basement membrane (GBM), a key component of the kidney's filtration barrier. GBM adopts undulated, complex shapes and becomes more irregular and bumpy in a diseased kidney, which represents a particular challenge for its automated segmentation and shape quantification. 
Segmentation accuracy of our model matches the agreement of expert annotators. We demonstrate that the continuity-aware training improves reconstruction smoothness and suppresses a periodic terracing artifact with minimal accuracy cost. We quantify GBM thickness in the reconstructed 3D anisotropic shapes and demonstrate the capacity of our model to faithfully capture GBM thickening in a diseased kidney. Our approach paves the way to fully automated anisotropic 3D morphometry of biological structures without dense volumetric labels or image restoration. The code is available at \mbox{\pdfstartlink attr{/Border[0 0 0]} user{/Subtype/Link/A<</Type/Action/S/URI/URI(https://github.com/bozeklab/gbm-seg/tree/main)>>}\url{https://github.com/bozeklab/gbm-seg/tree/main}\pdfendlink}. The dataset is available at \mbox{\pdfstartlink attr{/Border[0 0 0]} user{/Subtype/Link/A<</Type/Action/S/URI/URI(https://doi.org/10.5281/zenodo.21793752)>>}\url{https://doi.org/10.5281/zenodo.21793752}\pdfendlink}.
\end{abstract}

\begin{IEEEkeywords}
anisotropic volumetric segmentation, continuity loss, consensus inference, GPU morphometry, 3D reconstruction, deep learning, SwinUNETR, 3D U-Net, renal pathology, glomerular basement membrane
\end{IEEEkeywords}

\section{Introduction}

Tissue clearing techniques that incorporate swelling or expansion offer an unprecedented opportunity to characterize the
3D morphology and molecular organization of biological structures at
sub-diffraction resolution~\cite{unnersjo-jess_fast_2021, unnersjo-jess_super-resolution_2016, butt_molecular_2020}. By chemically
clearing and swelling fixed tissue, sub-diffraction limit resolution approaching 100~nm can
be achieved using standard confocal microscopes~\cite{unnersjo-jess_fast_2021}.
Since confocal microscopy is widely available, this technique makes quantification of the morphology of thin membranes, extracellular
matrices, and cytoskeletal networks accessible to broad research community. Furthermore, simplified
imaging protocols on commonly available equipment make this
approach potentially well suited for clinical translation~\cite{unnersjo-jess_fast_2021}. Transmission electron microscopy (TEM) and
focused ion beam scanning electron microscopy (FIB-SEM) achieve higher lateral
resolution, but require destructive sample preparation (heavy-metal staining,
resin embedding, and physical sectioning or irreversible ion-beam ablation)~\cite{miyaki_three-dimensional_2020}.

Despite this potential, quantitative analysis of such 3D images is constrained by the anisotropy
inherent to all single-objective fluorescence microscopy. Because the objective captures
light from only one direction, the 3D point spread function (PSF) is elongated along the
optical axis, giving more precise spatial information laterally than axially. Multi-objective
setups can generate isotropic images but are impractical for routine use. Moreover, acquiring
isotropic voxels via a point-scanning confocal microscope critically extends acquisition time
 from minutes to hours per stack. Typically, the voxel size is set to fulfill the Nyquist criterion for enough sampling to achieve the highest theoretical resolution. This anisotropy directly impedes automated 3D
segmentation~\cite{isensee_nnu-net_2021} and degrades reconstruction
accuracy of the imaged objects~\cite{ning_deep_2023, park_deep_2022}. For thin structures with high variation
along all three axes or whose thickness is comparable to the axial voxel pitch, insufficient
axial sampling causes discontinuities in the segmented volume that propagate into
downstream analysis. Importantly, training a segmentation model on a resampled
isotropic volume requires annotations in the upsampled space, critically increasing the
labeling burden.

In this work, we present a general computational pipeline that directly addresses
the problem of segmentation and reconstruction of biological structures imaged in 3D anisotropic stacks. To infer the missing 3D information we use the structural continuity in the lateral plane in which the resolution is high
and annotations are reliable. 
We exploit this structural continuity to enable
axially smooth segmentation by introducing a patch-level rotation augmentation strategy which consist of generating lateral cross-sections as surrogate axial views. Additionally, we introduce a
$z$-axis continuity loss that penalizes inter-slice discontinuities.
Together, these components enable 3D deep learning segmentation from
anisotropic volumes while training exclusively on annotations at the native
acquisition resolution, without dense upsampled labeling, image restoration, or
multi-objective imaging setups. 
To analyze the segmented structures we developed a fully GPU-accelerated morphometric module which computes
membrane thickness. It generates the measurements efficiently without skeletonization, mesh reconstruction, or structure-specific
geometric assumptions.

We apply our pipeline to one of
the most challenging and complex biological structures, the glomerular basement membrane (GBM).
The GBM is a thin extracellular membrane situated between the glomerular endothelium
and the podocyte foot processes and is a key component of the kidney's
filtration barrier~\cite{suleiman_nanoscale_2013}. It is an ideal proof-of-concept
structure for our pipeline for three reasons. First, it is only 200--400~nm thick. Resolving such a thin membrane requires
sub-diffraction imaging, and is especially challenging given the axial anisotropy.
Second, its surface is highly curved and varies rapidly in 3D which requires accurate segmentation and dedicated morphometry estimation methods. Third, structural alterations to the GBM are diagnostically important in several
kidney diseases and are the central hallmark of Alport syndrome, the second most
common genetic kidney disease~\cite{cosgrove_collagen_2017}. In current clinical
practice the GBM is evaluated from sparsely sampled 2D TEM cross-sections, which
introduce sectioning artifacts and sampling bias and discard the 3D morphology, limitations that contribute to the diagnostic difficulty of Alport syndrome, where
early GBM changes are subtle and easily missed. To the best of our knowledge,
existing automated GBM analysis methods are restricted to these 2D
cross-sections~\cite{rangayyan_segmentation_2010, wu_segmentation_2010,
cao_automatic_2019, wang_segmentation_2024}; the only 3D GBM morphometry reported to
date is that of Ali et~al.~\cite{ali_super-resolved_2025}, and no
fully automated pipeline exists at nanoscale resolution.

We validate the pipeline on a multi-channel high-resolution confocal dataset of
mouse glomeruli spanning three experimental groups: wild-type controls, a collagen~IV
mutation model of Alport syndrome, and a podocin mutation model of nephrotic syndrome type~2 (NPHS2). Images were prepared using validated
tissue clearing and swelling protocols~\cite{unnersjo-jess_fast_2021,
unnersjo-jess_super-resolution_2016}. The NPHS2 group, whose
GBM exhibits particularly high geometric variability, serves as the hardest test
case for our method.

The primary contributions of this work are:
\begin{itemize}
    \item A general deep learning pipeline for 3D segmentation of thin,
    morphometrically complex structures from anisotropic high-resolution confocal
    volumes, trained on native-grid annotations without dense upsampled labeling or
    image restoration.
    \item A continuity-aware training strategy that through patch rotation and continuity loss exploits the high-resolution lateral plane for improved axial plane reconstruction.
    \item Depth-aware adaptations of both a convolutional (3D U-Net) and a
    transformer (SwinUNETR) backbone.
    \item A fully GPU-accelerated morphometric analysis module
    producing a thickness surface map in 3D.
    \item The first automated 3D GBM reconstruction and morphometric analysis from
    high-resolution fluorescence microscopy.
\end{itemize}

\section{Related Work}

\subsection{Handling Anisotropy in Volumetric Segmentation}

The standard 3D U-Net~\cite{ourselin_3d_2016} and its transformer-based successors
such as Swin UNETR~\cite{crimi_swin_2022} apply spatially uniform operations:
a $3\times3\times3$ convolutional kernel or a uniform patch-tokenization scheme which treat
all three axes symmetrically, sampling equal numbers of voxels in each direction.
When voxel spacing is anisotropic, these operations are equally sized in 3D but act on unequal
physical volumes, which negatively affects segmentation performance~\cite{isensee_nnu-net_2021}.

The most common approach to address this is \emph{resampling} of volumes to an isotropic 
spacing before training of the segmentation model, so that the network is trained on voxels of cubic shape. nnU-Net
\cite{isensee_nnu-net_2021} automates this, using axis-specific percentile heuristics
and pseudo-2D convolutions in the early encoder layers to suppress artifacts along
the coarse axis. While effective, training on
isotropically resampled volumes carries the cost of additional annotations in that same upsampled
space. In the case of GBM, producing dense 3D labels for multi-channel confocal volumes is
laborious given the structural complexity and the expert anatomical knowledge
required. Producing annotations at 5--6$\times$ the number of $z$-slices makes the
task prohibitively large. 

Anisotropy-aware convolutions reshape kernels to the voxel geometry~\cite{crimi_automatic_2018}, but do not generate information along the under-sampled axial dimension. Connectivity-
preserving loss functions represent an additional solution: clDice~\cite{shit_cldice_2021}
penalizes skeleton breaks in tubular structures, and the persistent-homology loss of
Clough~et~al.~\cite{clough_topological_2022} encodes topological priors over connected
components and handles. These losses enforce a fixed, predefined topology, for example that the structure stays connected or has a set number of loops. This does not fit the variable shape of the GBM.

A parallel line of work restores axial resolution at the image level before
segmentation: supervised (CARE~\cite{weigert_content-aware_2018}) and self-supervised (IsoNet~\cite{descoteaux_isotropic_2017}, OT-CycleGAN~\cite{sim_optimal_2020}, Self-Net~\cite{ning_deep_2023}) methods demonstrate impressive quality of reconstructed isotropic volumes from
anisotropic fluorescence images. However, restoration improves the image, not the segmentation: the enlarged isotropic volumes still require manual annotation.

\subsection{3D Fluorescence Morphometry of the Glomerulus}

Prior automated  analysis of GBM morphology is uniquely 2D and TEM-based. Active contour and
skeletonization methods~\cite{rangayyan_segmentation_2010, wu_segmentation_2010} as well as random forest stacks~\cite{cao_automatic_2019} and transformer-based
RADS-Net~\cite{wang_segmentation_2024} were developed for such image segmentation. More
recently, foundation-model approaches such as GBMSeg~\cite{liu_feature-prompting_2024} achieved
strong one-shot segmentation, and TEM-AID~\cite{ma_ai-based_2025} demonstrated clinical-
grade performance. All these approaches operate in
2D where estimated thickness might vary depending on the
sectioning angle and where the full 3D surface geometry of the GBM is missing.

The most closely related work to ours is that of Ali et~al.~\cite{ali_super-resolved_2025}, who combined expansion microscopy with confocal imaging to reconstruct whole mouse glomeruli in 3D at nanoscale resolution. Rather than the sparse 2D cross-sections of conventional TEM, they recover the full glomerular surface and produce the first global GBM thickness maps of intact glomeruli. It relies, however, on manual correction and a technically demanding expansion protocol.

\section{Methodology}

\begin{figure*}[t]
  \centering
  \includegraphics[width=\textwidth]{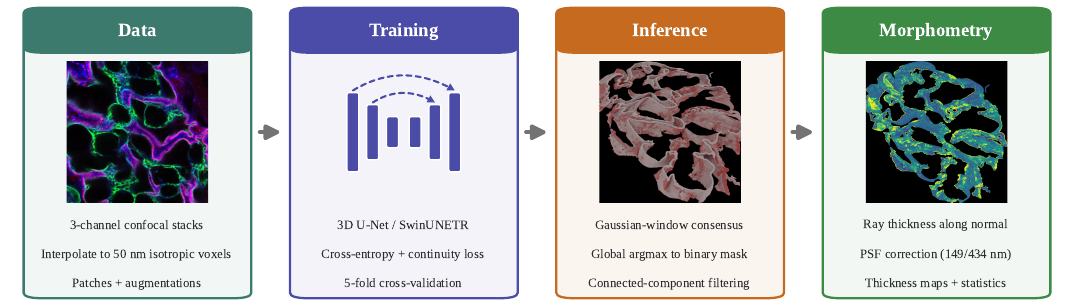}
  \caption{Overview of the pipeline, from anisotropic confocal input
  through continuity-aware training to consensus inference, post-processing, and
   morphometry.}
  \label{fig:pipeline}
\end{figure*}

The proposed framework addresses 3D segmentation and morphometric analysis of the
GBM from anisotropic confocal volumes. It is composed of a preprocessing stage, a
deep learning segmentation model trained with a continuity-aware objective, a
consensus-based inference procedure, a fully GPU-accelerated morphometric analysis
module, and a visualization pipeline. Fig.~\ref{fig:pipeline} provides a high-level
overview of the complete workflow.

\subsection{Data and Preprocessing}\label{sec:data}

\subsubsection{Dataset}

The primary dataset consists of confocal microscopy volumes of optically cleared
and swelled mouse kidney glomeruli acquired across three experimental groups:
wild-type controls, a collagen~IV mutation (Alport) model, and a podocin mutation (NPHS2) model. Each volume contains three
fluorescent channels, Nephrin (podocyte marker), Collagen~IV (which labels the basal aspect of the GBM), and WGA
(a glycoprotein stain that labels the full GBM and cell surfaces), encoded as 8-bit
intensity values. Volumes were acquired at $2048 \times 2048$ laterally with $12$--$16$ axial slices (most commonly $12$ or $14$), at an
anisotropic voxel pitch of approximately $47 \times 47 \times 300$~nm, giving a
lateral-to-axial spacing ratio of roughly 6:1. This lateral pitch oversamples the microscope's $\sim$150~nm lateral optical resolution
roughly threefold, as needed for deconvolution. Sampling more finely would increase acquisition time without adding any information about the GBM morphology.

In total, 12 volumes from 8 mice were manually annotated: 5 Alport, 4 NPHS2, and 3 control. A further 62 volumes were left unannotated: 15 Alport, 18 NPHS2, and 29 control. The GBM was delineated as the region between the collagen~IV signal at its basal, inner boundary and the nephrin signal at its outer boundary, with WGA used where either marker was locally absent. For evaluation, one sub-volume per group was independently annotated by three
experts, enabling inter-annotator agreement analysis. All annotations were made on the
original, non-interpolated volumes.

\subsubsection{Volumetric Interpolation}

Because the acquisition contains only $12$--$16$ axial slices while the lateral plane is
far more finely sampled, all volumes are resampled to a $50\times50\times50$~nm
\emph{isotropic} working grid. The lateral axes are resized to $50$~nm during dataset
preparation, and the depth axis (acquired at $\sim$300~nm) is upsampled by a factor
of six with trilinear interpolation prior to both training and inference. Isotropy is not
merely convenient for the network: it is required by the ray-based morphometry
(Section~\ref{sec:morph}), which measures thickness as a Euclidean distance along
arbitrary 3D directions. The three input channels (Nephrin, Collagen~IV, WGA) are interpolated in
PyTorch's built-in trilinear mode. Label
interpolation is not performed: each annotated $z$-slice is instead replicated across the
six interpolated positions. This is deliberate: interpolating binary segmentation masks
would introduce incorrect intermediate labels in regions of genuine $z$-discontinuity, and
replication restricts the annotation to the original acquisition volume.

\subsubsection{Patch Extraction}

Given the memory demands of full-volume 3D convolutions, all processing operates
on overlapping 3D patches of size $12 \times 256 \times 256$ (after interpolation)
extracted via a sliding window. Training uses a moderate stride
($8 \times 192 \times 192$) that limits redundant exposure to the heavily replicated
axial labels, whereas inference uses a dense stride (a single slice axially,
$64$~px laterally) so that every voxel is covered by many patches for the consensus
aggregation described in Section~\ref{sec:inference}.

\subsubsection{Data Augmentation}

Augmentation is applied at two levels. \emph{Entire volume-level} augmentation
operates on each whole annotated volume in voxel space. Applying it at every epoch would be prohibitively costly. We therefore precomputed a fixed set of geometric variants of each volume, as well as an isotropic zoom and a rotational twist around the depth axis, and cached them to disk. This expanded the training set.

\emph{Patch-level} augmentation is applied to each extracted patch at
runtime during training and is central to
resolving the axial-discontinuity problem. Its key component is random rotation around all three axes. The lateral plane is finely sampled and reliably annotated, whereas the axial plane is not. Rotating a patch brings well-resolved lateral structure into what would otherwise be the axial direction. This uses the lateral continuity as a surrogate for axial smoothness on the interpolated slices, for which no reliable ground truth exists. Although the replicated labels do not provide true
supervision along the interpolated depth, we found this augmentation to be key for eliminating axial
fragmentation, particularly in the geometrically complex NPHS2 volumes
(Section~\ref{sec:results}). Additional patch-level augmentations (Gaussian blurring, random cropping, and
random channel dropout) are each applied with a fixed probability to improve
robustness to imaging variation.

\subsection{Segmentation Model}

We evaluate two segmentation backbones under the same continuity-aware training
strategy, a custom 3D U-Net~\cite{ourselin_3d_2016} and a custom SwinUNETR~\cite{crimi_swin_2022}. Both are bespoke implementations that adapt the handling
of the depth axis to its small number of slices, which stays low even after interpolation, rather than treating all three axes
symmetrically, and both are trained and evaluated identically so that the comparison
isolates the effect of the backbone.

The 3D U-Net follows an encoder--decoder structure with skip connections. We modified the standard network to pool less aggressively along the $z$-axis: a standard
3D U-Net halves the resolution in all three axes at every stage, which would
collapse the shallow axial extent to fewer than two slices after only a few stages.
Instead, in our solution every encoder pool halves the lateral ($xy$) resolution but reduces the depth by
only a small fixed number of two slices. A $12\times256\times256$
patch therefore is reduced along $z$ from $12\to10\to8\to6$ in the four blocks of encoder while
the lateral dimensions are halved at each block. This way the feature maps are reduced to half of the input depth at the end of the encoder.

The SwinUNETR variant is re-engineered for shallow stacks through
three changes, each motivated by the small number of $z$-slices in our volumes. First,
\emph{patch merging} is altered along the z-axis. Each merging step halves the $xy$-plane but deducts only two slices from the
depth ($12\to10\to8\to6$). Second, since a 6--12-slice axis is too shallow for the hierarchical
$z$-tokenization, each stage attends over a window that
spans the \emph{full} $z$-dimension and the shifted-window cyclic shift is applied
in $xy$-dimensions only. Third, the patch-embedding uses a $z$-direction convolution with kernel size~5. Its field of view differs from the number of inserted slices (6), letting the model combine both replicated and original slices at different proportions (Section~\ref{sec:data}). The remaining hyperparameters
are as in the original method.


\subsection{Training}

\subsubsection{Loss Function}

We test two loss function combinations: a class-weighed
cross-entropy (CE) baseline, and CE combined with a
$z$-axis continuity term,

\begin{equation}
  \mathcal{L} = \alpha \, \mathcal{L}_{\mathrm{CE}} + \beta \, \mathcal{L}_{\mathrm{cont}},
  \label{eq:loss}
\end{equation}

\noindent where $\mathcal{L}_{\mathrm{CE}}$ is the class-weighed cross-entropy, with
per-class weights of 0.05 for background and 0.95 for the foreground GBM, $\mathcal{L}_{\mathrm{cont}}$ is the
continuity term, and $\alpha, \beta$ are \emph{fixed} scalar weights ($\alpha = 0.7$,
$\beta = 0.3$). The per-class weights and the $\alpha, \beta$ coefficients were chosen
experimentally.

The continuity term penalizes axial variation of the predicted class:

\begin{equation}
  \mathcal{L}_{\mathrm{cont}} = \frac{1}{(D-1)\,H\,W}
  \sum_{z=1}^{D-1} \sum_{y,x}
  \bigl| p_{z,y,x} - p_{z-1,y,x} \bigr|,
  \label{eq:cont}
\end{equation}

\noindent where $p_{z,y,x} = \mathrm{softmax}(\hat{f}_{z,y,x})$ is the foreground
probability obtained by a softmax over the class dimension of the logits $\hat{f}$, and
$D$, $H$, $W$ are the depth, height, and width of the patch. Operating on the softmax-derived probability rather than on raw logits keeps the continuity loss bounded and scale-invariant, which allows a fixed weight $\beta$. The term encourages axially
coherent predictions on the interpolated slices for which no reliable ground truth
exists. As reported in Section~\ref{sec:results}, adding this term leaves segmentation accuracy
statistically unchanged while measurably improving the axial smoothness of the
reconstruction; it therefore acts as a coherence regularizer rather than an
accuracy-driven objective.

\subsubsection{Optimization}

The network is optimized with Adam~\cite{kingma_adam_2014} under a polynomial-decay learning-rate schedule
(power~0.9)~\cite{chen_deeplab_2018}, using mixed-precision training~\cite{micikevicius_mixed_2017} distributed across multiple A100 GPUs. Because
each volume yields a large number of highly overlapping patches and the $z$-depth is
limited even after interpolation, the model is susceptible to overfitting on
axial-direction patterns. Validation metrics are therefore evaluated after a fixed
number of samples rather than at epoch boundaries, providing more frequent feedback
during training.

\subsection{Inference}\label{sec:inference}

Inference uses a consensus-based patch aggregation strategy. The trained network is applied to densely overlapping patches of the interpolated input volume, with an axial stride of a single slice and a lateral stride of 64 voxels. Predicted probabilities of voxels in each patch are weighed by a 3D Gaussian window that
down-weights patch edges, where predictions are least reliable, and added into a
probability tensor of the full interpolated volume. A single global argmax
over the accumulated probabilities produces the binary segmentation mask.

This weighed-consensus strategy improves boundary coherence and reduces patch-edge
artifacts compared with independent per-patch argmax or uniform logit summation,
particularly in regions where the GBM crosses patch boundaries at oblique angles. The
mask is retained on the interpolated working volume, on which all subsequent
post-processing and morphometry operate.

Post-processing includes two rounds of connected-component filtering: first in 2D on each
$z$-slice independently to remove small isolated regions, followed by a 3D
connected-component analysis to eliminate small spurious 3D volumes. The
minimum-size thresholds for both filters are configurable at inference time.

\subsection{Axial-Continuity Metrics}\label{sec:continuity_metrics}

We evaluate the axial smoothness of the reconstructed volumes with four metrics.
Three measure the along-$z$ roughness of the mask: the total variation of the mask
along $z$ normalized by foreground volume (Z-TV per foreground voxel), the mean number
of $0\!\leftrightarrow\!1$ transitions per foreground column along $z$ (a single
membrane crossing yields two), and the mean intersection-over-union between adjacent
$z$-slices. All are reported alongside the size of the foreground region, so that an
output is not judged ``smoother'' simply because it contains less membrane.

The fourth metric, the \emph{terracing index}, targets the specific stair-step
artifact of the $6\times$ slice replication (Section~\ref{sec:data}). For each pair of
adjacent $z$-slices we count how many voxels flip between them. A smooth membrane
changes gradually, so these counts are similar from slice to slice; a terraced one
instead leaves them near zero inside each replicated block and spikes at every sixth
slice, where the native slices meet. To capture this, we sort the slice gaps into six
groups by their position in the repeating six-slice cycle and compare each group with
the overall average. We use $D_i$ to denote the number of voxels that flip between
interpolated slices $i$ and $i+1$, $\bar{D}$ - the average of $D_i$ over all $Z-1$
gaps, and $\bar{D}_k$ - the average over the gaps at cycle position $k$ (those with
$i \bmod 6 = k$),
\begin{equation}
  \mathrm{TI} = \max_{0 \le k < 6} \frac{\bar{D}_k}{\bar{D}}.
  \label{eq:terracing}
\end{equation}
\noindent The index equals $1$ when the changes are spread evenly across slices
(smooth) and rises toward $6$ when they all fall on the block boundaries (fully
terraced). It is a measure of this replication artifact that was designed for the
purpose of this study, rather than a standard metric.

\subsection{Morphometric Analysis}\label{sec:morph}

\begin{figure}[t]
  \centering
  \includegraphics[width=\columnwidth]{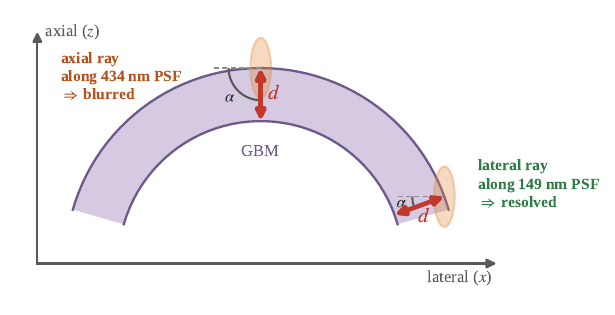}
  \caption{GPU thickness measurement and its orientation dependence, shown as an
  $x$--$z$ cross-section. A ray cast normal to the membrane gives the raw thickness;
  the anisotropic PSF blurs axial rays more than lateral ones, which the PSF
  correction removes.}
  \label{fig:morphometry}
\end{figure}

All morphometric computations are implemented in PyTorch and executed on the GPU,
operating directly on the voxel mask without mesh reconstruction or skeletonization.
The measurement is intrinsically 3D and orientation-dependent; its geometry
and the anisotropic-PSF correction are illustrated in
Fig.~\ref{fig:morphometry}.

\subsubsection{Surface Detection}

Surface voxels are identified by convolving the binary segmentation mask with a
$3 \times 3 \times 3$ kernel that counts each voxel's 26 neighbors. A foreground
voxel with fewer than 26 foreground neighbors, and therefore at least one
background neighbor, is classified as a surface voxel, forming a binary surface
mask $\mathcal{S}$.

\subsubsection{Surface Normal Estimation}

For each surface voxel $v \in \mathcal{S}$, directional gradients are computed along
the $x$, $y$, and $z$ axes using three separable 3D convolution kernels. The resulting
gradient vectors:
\begin{equation}
  \mathbf{s}(v) = \bigl( s_x(v),\; s_y(v),\; s_z(v) \bigr)
\end{equation}
are smoothed by averaging over the surface voxels within a configurable $5^3$
neighborhood kernel to reduce voxel-level noise and enforce local smoothness of the
estimated surface normal field.

\subsubsection{Inward Normal Direction Selection}

The smoothed gradient vector at each surface voxel is compared, via dot product, with the
26 directions that point from that voxel to its neighbors in the surrounding
$3 \times 3 \times 3$ cube. These 26 neighbors are the 6 that share a face with the
center voxel, the 12 that share an edge, and the 8 that share a corner. The direction
$\hat{\mathbf{n}}(v)$ with the largest dot product with $\mathbf{s}(v)$ is selected as the
inward-pointing probing direction for that voxel.

\subsubsection{Ray-Based Thickness Measurement}

A ray is cast from each surface voxel $v$ in direction $\hat{\mathbf{n}}(v)$ using
parametric 3D line equations. The intersection with the opposite surface is found as
the nearest surface voxel $v'$ encountered along the ray, and the raw measured
thickness is:
\begin{equation}
  d(v) = \min_{v' \in \mathcal{S}} \bigl\| \mathbf{p}(v) - \mathbf{p}(v') \bigr\|
\end{equation}
\noindent where $\mathbf{p}(v)$ denotes the physical coordinate of voxel $v$ in
nanometres, obtained by scaling voxel indices by the isotropic $50$~nm voxel size
(Section~\ref{sec:data}).

\subsubsection{PSF-Corrected Thickness}

Raw ray distances are affected by the anisotropic PSF of the confocal microscope: a
measurement taken along a direction with angle $\alpha$ relative to the $xy$-plane
probes a mixture of lateral and axial resolution. Approximating the PSF as a 3D Gaussian
ellipsoid, its variance along the probing direction is
$\mathrm{PSF}_{\mathrm{lat}}^2 \cos^2\!\alpha + \mathrm{PSF}_{\mathrm{ax}}^2 \sin^2\!\alpha$;
we subtract this contribution from the squared apparent thickness and take the square root:
\begin{equation}
  T_{\mathrm{true}} = \sqrt{\, d^2 - \bigl(\mathrm{PSF}_{\mathrm{lat}}^2 \cos^2\!\alpha
  + \mathrm{PSF}_{\mathrm{ax}}^2 \sin^2\!\alpha \bigr) \,}
  \label{eq:psf}
\end{equation}
\noindent where $\mathrm{PSF}_{\mathrm{lat}}$ and $\mathrm{PSF}_{\mathrm{ax}}$ are
the lateral and axial full-width-at-half-maximum values of the microscope's PSF
($\mathrm{PSF}_{\mathrm{lat}} = 149$~nm and $\mathrm{PSF}_{\mathrm{ax}} = 434$~nm
for our confocal acquisitions), and $\alpha$ is the angle between the probing direction
and the lateral plane, derived directly from $\hat{\mathbf{n}}(v)$. At some surface points the argument of the square root is negative, meaning the raw measured distance is smaller than the PSF contribution, most likely where imperfect segmentation locally underestimates the membrane thickness. These points are set to zero and excluded from the thickness statistics. In the remainder of the paper, \emph{thickness} refers to this PSF-corrected
value unless stated otherwise.

\subsection{Statistical Analysis}\label{sec:stats}

Several glomeruli are imaged per animal. Although these are distinct structures, they share the animal's genotype, age, and tissue preparation and are therefore not independent (pseudo-replicates). The disease group is a property of the animal, and because disease severity varies between glomeruli within an animal, the per-animal aggregate is its most representative measure. The animal is therefore the unit of replication for group comparisons: we aggregate to one value per animal, and include image-level analysis for transparency only.

GBM thickness values are right-skewed: most are small, with a long tail of larger
values. This skew reflects our dense, whole-surface measurement, which captures local
thickenings that sparse baseline-section measurements omit.

Group differences are assessed with a single omnibus Kruskal--Wallis test on the per-mouse
mean thickness (the biological replicate; $n = 4$, $3$, and $2$ animals for the control,
NPHS2, and Alport groups). Because every group has fewer than five animals, we
report the \emph{exact} permutation $p$, enumerated over all $1260$ group-label
assignments, rather than the chi-square approximation, which is invalid at this size. Given
the small cohort we foreground the per-group means and Cliff's~$\delta$ effect sizes as the
primary evidence and treat the omnibus $p$ as corroboration.

\subsection{Visualization}

For 3D rendering, the binary segmentation mask is converted to a
surface mesh using the marching cubes algorithm. The vertex coordinates, face
connectivity, and per-vertex thickness values are rendered in Blender with a continuous
color scale mapped to thickness, so that spatial variation in GBM thickness is directly
visible on the reconstructed surface geometry.

In addition, the per-vertex thickness is projected top-down along the $z$-axis to a 2D
heatmap, enabling rapid qualitative assessment of thickness patterns and standardized
cross-sample comparison without 3D rendering software.

\section{Results}\label{sec:results}

Unless stated otherwise, segmentation accuracy is reported as five-fold
\emph{subject-wise} cross-validation Dice on the foreground (GBM) class. Folds contain separate mouse data, so no animal appears in both training and validation.
Because every configuration is trained and evaluated on the same fold splits, two
configurations can be compared on each fold as a matched pair; this removes the
fold-to-fold variation and makes small differences between configurations easier to detect. A
further set of $62$ unannotated whole-glomerulus volumes sampled from the same animals as the annotated data but representing disjoint areas of the GBM (Section~\ref{sec:discussion}), is used for the continuity and morphometric analyses.

\subsection{Segmentation Performance}\label{sec:segperf}

\subsubsection{Architecture and Loss}
Table~\ref{tab:cv} reports cross-validated Dice for the two backbones under both loss
functions. Two effects are clear. First, \emph{architecture matters}: 
SwinUNETR outperforms the 3D U-Net by roughly $0.09$ Dice ($\approx0.69$ vs.\
$\approx0.60$), a difference an order of magnitude larger than any other factor and well outside the fold-to-fold standard deviation. Second, \emph{the loss function does not affect Dice}: the paired
difference between the continuity and cross-entropy objectives is $-0.005 \pm 0.012$ for
SwinUNETR and $+0.006 \pm 0.020$ for the U-Net, smaller than its own standard deviation,
with the sign reversing between architectures. Segmentation accuracy is therefore
governed by the backbone, not by the continuity term, which motivates the dedicated
axial-coherence evaluation in Section~\ref{sec:continuity}. Given its clear accuracy
advantage, we adopt the SwinUNETR backbone for all subsequent results (expert
comparison, axial continuity, reconstruction, and morphometry).

\begin{table}[t]
\centering
\caption{Five-fold subject-wise cross-validation Dice (mean $\pm$ sd over folds).}
\label{tab:cv}
\begin{tabular}{llc}
\hline
Backbone & Loss & Dice \\
\hline
3D U-Net & Cross-entropy & $0.598 \pm 0.069$ \\
3D U-Net & Continuity & $0.604 \pm 0.051$ \\
SwinUNETR & Cross-entropy & $0.693 \pm 0.051$ \\
SwinUNETR & Continuity & $0.688 \pm 0.054$ \\
\hline
\end{tabular}
\end{table}

\subsubsection{Comparison to Expert Annotators}
\begin{figure}[t]
  \centering
  \includegraphics[width=\columnwidth]{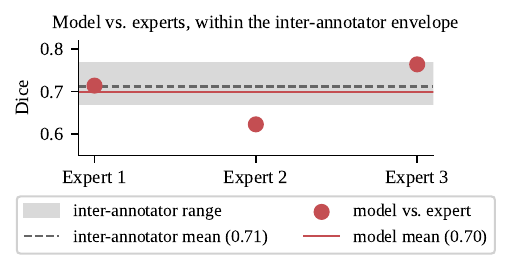}
  \caption{The SwinUNETR model agrees with each expert to a similar extent as the experts
  agree with one another.}
  \label{fig:expert_envelope}
\end{figure}

A subset of three volumes, one per group, was each independently annotated by three experts. We next compared the 
SwinUNETR model against every annotator and against the inter-annotator
agreement. As shown in Table~\ref{tab:expert} and
Fig.~\ref{fig:expert_envelope}, the model reaches a mean Dice of $0.700$ on data annotated by each of the
three experts. This is statistically indistinguishable from the mean inter-annotator agreement of $0.712$ and lies within the inter-annotator range ($0.67$--$0.77$). Thus, the model agrees
with the experts about as well as the experts agree with one another. Cross-entropy
reaches essentially the same mean ($0.701$ versus $0.700$).

\begin{table}[t]
\centering
\caption{Segmentation agreement on the labelled test crops, model versus experts and between experts.}
\label{tab:expert}
\begin{tabular}{lc}
\hline
Comparison & Dice \\
\hline
Model vs.\ Expert 1 & $0.714$ \\
Model vs.\ Expert 2 & $0.623$ \\
Model vs.\ Expert 3 & $0.764$ \\
\textbf{Model vs.\ experts (mean)} & $\mathbf{0.700}$ \\
\hline
Expert 1 vs.\ Expert 2 & $0.670$ \\
Expert 1 vs.\ Expert 3 & $0.766$ \\
Expert 2 vs.\ Expert 3 & $0.701$ \\
\textbf{Inter-annotator (mean)} & $\mathbf{0.712}$ \\
\hline
\end{tabular}
\end{table}

The NPHS2 group represents the most challenging data to segment, due to the high geometric variability
of the GBM and the resulting axial discontinuities even in the expert
annotations. In this group the continuity-aware training strategy is necessary to obtain coherent
segmentation, as quantified and illustrated in the next section
(Section~\ref{sec:continuity} and Fig.~\ref{fig:rotation}).

\subsection{Axial Continuity}\label{sec:continuity}

The cross-validation Dice above is calculated on the native (annotated)
$z$-slices and does not capture the axial continuity of the interpolated slices between them. We
therefore additionally evaluate axial smoothness on the reconstructed test volumes with the four
axial-continuity metrics of Section~\ref{sec:continuity_metrics}, each reported alongside the size
of the foreground region.

\subsubsection{Patch-Level Rotation}
Patch-level rotation is our primary mechanism to enhance axial coherence. Because the labels are
upsampled $6\times$ along $z$ by slice replication (Section~\ref{sec:data}), a model trained with
weak rotation could memorize this period-6 structure, producing a step-like membrane rather than a continuous sheet. Rotating patches during
training breaks this axis-aligned periodicity and suppresses this artifact.

\begin{figure}[t]
  \centering
  \includegraphics[width=\columnwidth]{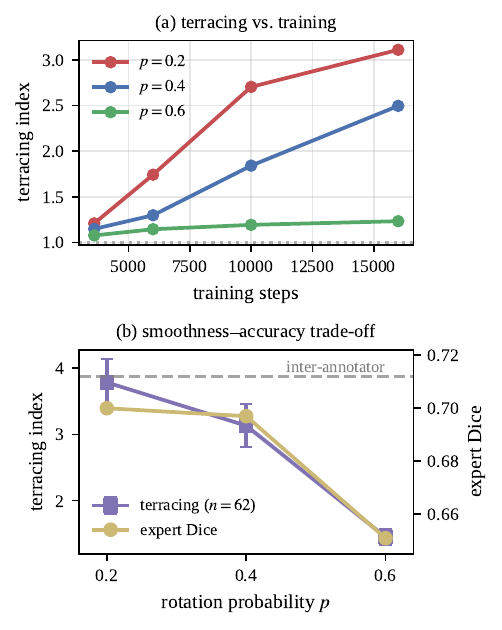}
  \caption{Rotation augmentation suppresses period-6 axial terracing, trading a small
  amount of segmentation accuracy for smoother reconstruction as the rotation
  probability rises. The example reconstruction is an NPHS2 volume, the most
  challenging group.}
  \label{fig:rotation}
\end{figure}

We quantify the effect with the \emph{terracing index} (Section~\ref{sec:continuity_metrics}),
which ranges from $1$ (smooth, transitions spread evenly across slices) to $6$ (fully terraced,
all transitions at the block boundaries).
We tested the rotation probability $p \in \{0.2, 0.4, 0.6\}$ at a fixed training budget and
measured the terracing index over all $62$ test volumes (Table~\ref{tab:rotation},
Fig.~\ref{fig:rotation}) and found that the step-artifact is suppressed monotonically, from
$4.05\pm0.30$ without rotation, through $3.79\pm0.35$ at $p=0.2$ and $3.13\pm0.32$ at
$p=0.4$, to $1.45\pm0.12$ at $p=0.6$. The reduction is strongly nonlinear: moderate
rotation ($p{\le}0.4$) shows only a mild effect on the smoothness, and a substantial effect appears
only at $p=0.6$. Per-checkpoint segmentation of a representative volume
(Fig.~\ref{fig:rotation}a) further shows that terracing develops progressively over
training, so stronger rotation is needed to diminish it throughout the training.

This continuity requirement trades off against accuracy. Agreement with the three expert
annotators (Section~\ref{sec:segperf}) is essentially unchanged at $p=0.4$ (Dice $0.697$ versus $0.700$
at $p=0.2$) but falls to $0.651$ at $p=0.6$, just below the inter-annotator agreement of
$0.712$ (Fig.~\ref{fig:rotation}b). The size of foreground is stable across the tests, so
the smoother surface is not a result of predicting less membrane. The choice of $p$ is therefore a trade-off. A lower value ($p=0.4$) preserves segmentation accuracy but leaves most of the terracing intact. A higher value ($p=0.6$) yields continuous, largely terracing-free reconstructions, as required for morphometry and visualization, at a slightly lower segmentation accuracy (Section~\ref{sec:discussion}).

\begin{table}[t]
\centering
\caption{Rotation augmentation versus axial terracing and segmentation accuracy across the rotation-probability sweep. The terracing index is mean $\pm$ sd.}
\label{tab:rotation}
\begin{tabular}{lcccc}
\hline
Rotation $p$ & Terracing $\downarrow$ & Expert Dice & FG voxels & FG\% \\
\hline
0.2 & 3.79$\pm$0.35 & 0.700 & 12.76M & 3.24 \\
0.4 & 3.13$\pm$0.32 & 0.697 & 12.80M & 3.25 \\
0.6 & \textbf{1.45$\pm$0.12} & \textbf{0.651} & 13.65M & 3.47 \\
\hline
\end{tabular}
\end{table}

\subsubsection{Continuity Loss}
In addition to rotation augmentation, the continuity loss provides a further
improvement in the volume smoothness. Table~\ref{tab:continuity} compares the
SwinUNETR model trained with cross-entropy against the same model trained with the
continuity objective, on the test set. All three continuity metrics improve with the continuity loss. Z-TV per foreground voxel and per-column flip count are each lower by $\approx7\%$, and adjacent-slice IoU is higher, while the foreground size remains unchanged ($0.060$). The differences are
statistically significant (Welch $|z| \approx 3$--$5$ over the $62$ test volumes). Combined
with the tied Dice of Section~\ref{sec:segperf}, this shows that the continuity loss acts as a
\emph{coherence regularizer}: it improves the axial quality of the 3D reconstruction at no cost to segmentation
accuracy.

\begin{table}[t]
\centering
\caption{Axial continuity of the SwinUNETR reconstructions on the test set ($n=62$ volumes), continuity loss versus cross-entropy; values are mean $\pm$ sd.}
\label{tab:continuity}
\begin{tabular}{lccc}
\hline
Metric & Cross-entropy & Continuity & $\Delta$ \\
\hline
Z-TV per fg voxel $\downarrow$ & $0.145 \pm 0.019$ & $0.134 \pm 0.019$ & $-7.2\%$ \\
Flips per fg column $\downarrow$ & $2.605 \pm 0.229$ & $2.415 \pm 0.223$ & $-7.3\%$ \\
Adjacent-slice IoU $\uparrow$ & $0.876 \pm 0.013$ & $0.884 \pm 0.013$ & $+0.9\%$ \\
Foreground fraction & $0.060$ & $0.060$ & --- \\
\hline
\end{tabular}
\end{table}

\subsection{3D Reconstruction}

\begin{figure*}[t]
  \centering
  \includegraphics[width=\textwidth]{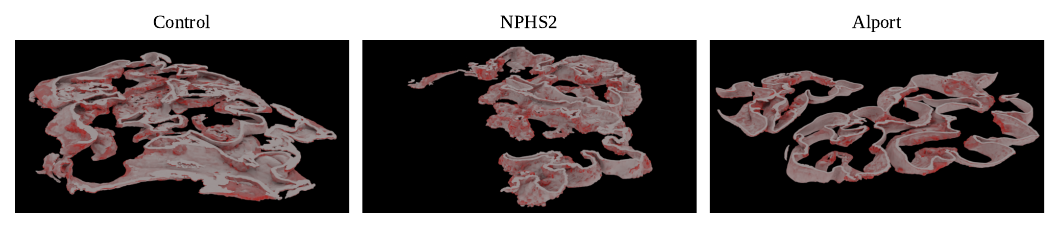}
  \caption{3D reconstructions of the GBM colored by membrane thickness, one
  representative specimen per group.}
  \label{fig:reconstruction}
\end{figure*}

Fig.~\ref{fig:reconstruction} shows reconstructions of representative specimens. The
marching-cubes mesh faithfully captures the highly convoluted GBM geometry,
including the curvature at capillary-loop bifurcations and the irregular surface of the
NPHS2 specimens. Consistent with the continuity results, the masks are
topologically intact and require none of the per-slice mesh repair that axial
fragmentation would otherwise necessitate.

\begin{figure*}[t]
  \centering
  \includegraphics[width=\textwidth]{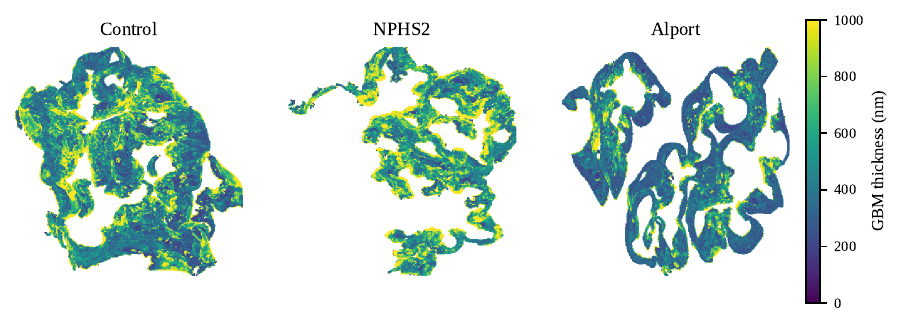}
  \caption{Top-down GBM thickness maps for one representative glomerulus
  per group on a shared thickness scale.}
  \label{fig:morphdemo}
\end{figure*}

\subsection{Morphometric Analysis}

Using the GPU morphometric module we produce membrane thickness across the full GBM
surface for every reconstructed test volume and calculate per-group thickness estimates (Section~\ref{sec:stats}). Table~\ref{tab:thickness} reports per-group means. The NPHS2 GBM is markedly thickened
($461 \pm 28$~nm) relative to both control ($386 \pm 4$~nm) and the
Alport group ($362 \pm 22$~nm). This separation is large and consistent (Cliff's~$\delta = 1$ against each group). Control and Alport differ by only $23$~nm;
although the two Alport animals fall below all four control animals, their image-level
distributions overlap and this small difference is not significant at $n = 2$. A single exact permutation Kruskal--Wallis test on the per-mouse means confirms an overall group difference ($H = 7.0$, exact $p = 6/1260 = 0.005$). However, this is the smallest $p$ a $4/3/2$ design can produce, and it reflects the complete rank ordering of the nine animals rather than the magnitude of the effect. The per-group means and effect sizes therefore carry the biological result. The choice of loss does not change this: the continuity and
cross-entropy models give near-identical thickness (Table~\ref{tab:thickness}), just as they gave near-identical Dice (Section~\ref{sec:segperf}). The group
separation is likewise unaffected by the amount of patch rotation: at $p{=}0.6$ the model
estimates the same NPHS2--control thickness difference (Table~\ref{tab:thickness}),
differing only by a small uniform $+12$--$18$~nm offset consistent with its marginally
larger foreground. 

\begin{table}[t]
\centering
\caption{GBM thickness per group (nm, mean $\pm$ sd) for the two losses and two rotation settings.}
\label{tab:thickness}
\begin{tabular}{lcccc}
\hline
Group & $n$ mice & Continuity & Continuity & Cross-entropy \\
 & & rotation $p{=}0.2$ & rotation $p{=}0.6$ & rotation $p{=}0.2$ \\
\hline
Control & 4 & $386 \pm 4$ & $402 \pm 10$ & $384 \pm 8$ \\
\textbf{NPHS2} & 3 & $\mathbf{461 \pm 28}$ & $\mathbf{473 \pm 22}$ & $\mathbf{458 \pm 27}$ \\
Alport & 2 & $362 \pm 22$ & $380 \pm 24$ & $367 \pm 22$ \\
\hline
\end{tabular}
\end{table}

Additionally, a top-down thickness map illustrates within-glomerulus spatial heterogeneity (Fig.~\ref{fig:morphdemo}). We show one representative glomerulus
per group on a shared thickness scale. The NPHS2 membrane appears visibly thicker across
the surface, consistent with the per-mouse comparison of Table~\ref{tab:thickness}, while
the maps also expose local variation within each individual
membrane.

\section{Discussion}\label{sec:discussion}

We have presented an end-to-end pipeline for automated 3D segmentation and morphometry of
the GBM from anisotropic confocal volumes, trained entirely on native-resolution
annotations. Three points merit emphasis.

First, our pipeline is \emph{not specific to the GBM}. It uses the well-resolved lateral
structure to guide axial smoothness where no axial ground truth exists, so it applies to
any thin structure imaged with strong axial anisotropy. The GBM is simply a demanding
test case.

Second, in our experiments, accuracy and axial coherence proved to be \emph{independent of one another}, determined by different parts of the method. The transformer backbone determines the Dice score, whereas the continuity-aware training determines the smoothness of the reconstruction. The Dice metric is evaluated only on the annotated slices, not the interpolated ones where terracing appears. It also does not capture continuity and can lead to inaccurate downstream morphometry. To
measure this effect we developed dedicated axial-continuity metrics.

Third, our pipeline faithfully captures different GBM morphologies. We observed the NPHS2 thickening
of the membrane which was
reproduced across losses and augmentation strategies. Control and Alport groups
remained indistinguishable at this sample size. This is itself informative: whole-membrane averages or sparse sampling are insufficiently sensitive to capture the subtle changes caused by the Alport syndrome. This motivates the dense, spatially-resolved 3D morphometry our pipeline enables.

There are several limitations of our study. The biological sample size is small, with only a few
animals per group, so the comparison of the morphometry among the groups should be read as preliminary
rather than definitive. The labeled test set comprises a single sub-volume per group annotated
by three experts, nine annotations in total. While this enabled us to calculate the 
inter-annotator variability, every Dice comparison against
the experts rests on this small set. In particular, the rotation test of
Section~\ref{sec:continuity}, where the $p{=}0.6$ setting decreases the performance of our model just below the
inter-annotator variability, should be read as indicative of the smoothness--accuracy
trade-off rather than a definitive accuracy ranking. To improve the robustness of these estimated would require
additional expert-annotated volumes. The axial-continuity and
terracing metrics, by contrast, are reported over all $62$ test volumes and do not share
this limitation. Moreover the $62$-volume analysis set and the
expert sub-volumes are drawn from the same animals as the training data, every animal but one
control, so the continuity, morphometric, and expert-agreement analyses gauge measurement
consistency within known subjects rather than generalization to different animals. Generalization to different animals is quantified only by the subject-wise cross-validation of
Section~\ref{sec:segperf}. 
The
backbone-internal choices in the SwinUNETR variant, namely the two-slice depth deduction,
the full-$z$ attention window, and the stem $z$-convolution, are motivated by the shallow axial extent and the axial terracing artifact but are not individually
ablated: our ablations vary the loss and the rotation probability, not the backbone, so
their separate contributions remain unquantified. 
The per-class cross-entropy weights and
the $\alpha, \beta$ loss coefficients were chosen experimentally; we did not perform a
systematic hyper-parameter search, so these values are not guaranteed to be optimal.
A rigorous morphometric analysis would also require a 3D region-of-interest step that
separates the capillary-loop GBM from mesangial cells that can be mis-segmented as
membrane; the present pipeline does not include such a step, so any residual mesangial
mis-segmentation is not excluded from the thickness statistics.
Finally, the pipeline is validated on
murine tissue; translation to human biopsy material, the ultimate diagnostic target,
remains future work, though the clearing-and-swelling protocol is compatible with same-day
imaging~\cite{unnersjo-jess_fast_2021}, supporting that translatability.

\section{Conclusion}

We introduced a general strategy for 3D deep-learning segmentation of thin,
morphometrically complex structures from anisotropic fluorescence volumes. We used lateral
structural continuity as a surrogate supervisor for axial smoothness and obviated the need for
dense volumetric annotation, image restoration, or multi-objective imaging setups. Applied to the GBM, it
enables, to our knowledge, the first automated 3D reconstruction and morphometric
analysis of this structure from confocal microscopy. It resolves the full spatial
distribution of GBM thickness rather than a single average. A single average leaves control and Alport membranes indistinguishable and easily misses subtle disease-related changes. The anisotropic
transformer backbone is the stronger segmenter; the continuity-aware objective, evaluated
with a dedicated axial-coherence metric, improves reconstruction smoothness at no accuracy
cost. Extending the approach to human biopsy tissue and to other thin, anisotropically
imaged structures is a natural next step.

\printbibliography

\end{document}